\documentclass[10pt, conference,a4paper]{IEEEtran}
\ifCLASSINFOpdf
\else
\fi
\usepackage{times}
\usepackage{epsfig}
\usepackage{graphicx}
\usepackage{amsmath}
\usepackage{amssymb}
\usepackage{multirow}
\usepackage{ulem}
\usepackage{url}
\usepackage[dvipsnames]{xcolor}
\usepackage{balance}

\begin{document}
%
\title{Answer-checking in Context: A Multi-modal Fully Attention Network for Visual Question Answering}

\author{\IEEEauthorblockN{Hantao Huang, Tao Han, Wei Han and Deep Yap }
\IEEEauthorblockA{
MediaTek, Singapore \\
Email: hantao.huang@mediatek.com}
\and
\IEEEauthorblockN{Cheng-Ming Chiang }
\IEEEauthorblockA{
MediaTek, Taiwan \\
Email: jimmy.chiang@mediatek.com}}

\maketitle
\begin{abstract}

Visual Question Answering (VQA) is challenging due to the complex cross-modal relations. It has received extensive attention from the research community. From the human perspective, to answer a visual question, one needs to read the question and then refer to the image to generate an answer. This answer will then be checked against the question and image again for the final confirmation. In this paper, we  mimic this process and propose a fully attention based VQA architecture. 
Moreover, an answer-checking module is proposed to perform a unified attention on the jointly answer, question and image representation to update the answer. This mimics the human answer checking process to consider the answer in the context. With answer-checking modules and transferred BERT layers, our model achieves the state-of-the-art accuracy 71.57\% using fewer parameters on VQA-v2.0 test-standard split.

\end{abstract}

\section{Introduction}

Visual Question Answering (VQA) is a popular multi-modal problem, which requires the machine to infer the correct answer from a given image and a related question. It receives extensive attention from the research community since the multi-modal fusion method could be extended to many more vision-language applications such as image grounding and image captioning.   However, it is still challenging to design an effective multi-modal fusion network.


Recently, attention mechanism becomes one of the most influential ideas in the deep learning research field. 
A few state-of-the-art results have been produced using attention based models, such as Transformer for machine translation \cite{ref-transformer} and BERT for natural language processing tasks  \cite{ref-bert}.  Attention is also one key idea for VQA architecture with many new attention networks proposed. For example, the dynamic intra-inter-modal attention is proposed to model the question-to-image attention as well as the image-to-question attention \cite{ref-dynamic_fusion}. 
The previous work uses separate attentions to perform the intra attention (e.g. question self-attention) and inter attention (eg. question-to-image attention).  However, our work proposes a unified attention flow to perform the multi-modal attention concurrently.  Our multi-modal attention is performed at the "answer checking" module by the answer-image-question self-attention.
We find that an answer related attention could mimic the answer checking process of human VQA to further improve the performance. For example, with only 1 layer of "answer checking" module, the VQA accuracy increases significantly from 60.24\% to 66.65\%. As such, we propose a multi-modal attention network to model the answer-question-image checking process. 

\begin{figure}[t]
	\centering 
	\includegraphics[width=0.48\textwidth]{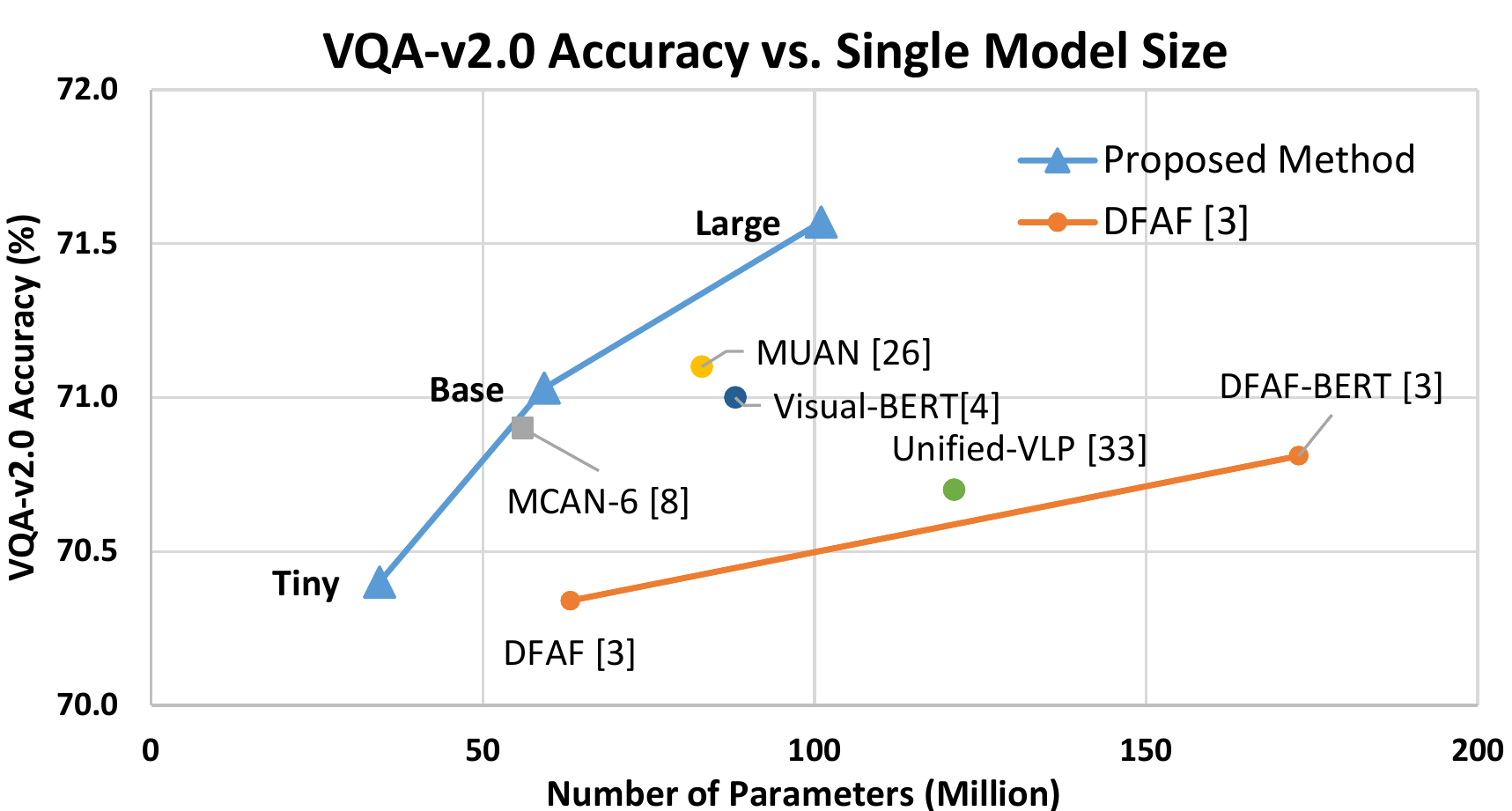}
	\caption{Proposed model performance (model size and accuracy) comparison with existing works. Our large model achieves better accuracy with smaller model size compared to the current state-of-the-art models.
	}
	\label{fig:intro1}
\end{figure}

In this paper, we propose an Attention based Read, Answer and Check (ARAC) architecture for VQA. 
Fig. \ref{fig:intro1} shows the performance comparisons on model size and accuracy with existing works.  We develop three models with different sizes. Compared to DFAF-BERT \cite{ref-dynamic_fusion}, we have a better utilization of the model size (number of parameters) to improve the accuracy. Note that since most state-of-the-art models such as Visual-BERT \cite{li2019visualbert} and VilBERT \cite{vilbert} are transformer or BERT based, the number of parameters is used to indicate the computation complexity.  
Our large model achieves a better accuracy yet smaller model size compared to the state-of-the-art models such as MLI-BERT \cite{gao2019multi} and VilBERT \cite{vilbert}.

Our work differentiates from the existing work in three manifolds. First, our architecture is much simpler without recurrent neural network such as LSTM. Second, we merge the intra-modal attention and inter-modal attention to perform one unified multi-modal attention. Last and most importantly, the answer-checking in context is new and first proposed to recursively update the current answer, question and image representation concurrently by adopting the proposed unified multi-modal attention.   
The contribution of this paper is summarized as follows. We propose


\begin{itemize}
\item 
A fully attention based VQA architecture: This is a novel parallel architecture without recurrent neural network. It uses three attention based modules to mimic the human behaviour (reading, answering and checking) to answer a question given an image. 

\item A multi-modal answer related attention flow
: 
We perform a multi-modal self-attention on the unified representation of the answer, question and image to check our initial answer representation.
The answer-checking process will update the answer representation based on the context of current answer, question and image representation. 
This mimics the human answer checking process to consider the answer in context. 

\item A layer-wise transfer learning: We investigate transferring different number of layers of pre-trainined model BERT \cite{ref-bert} to our VQA architecture.   
By transferring 6 layers of the pre-trained language model BERT and utilizing the answer-checking process, our single model achieves the state-of-the-art accuracy 71.57\% compared to the existing works on VQA-v2.0 test-standard split.
\end{itemize}

\section{Related Works}

\subsection{Visual Question Answering}

Visual question answering generates an answer based on the information provided by the image and question. This requires a good understanding of the question. 
Based on the question, the related information should be found from the image. As such, multi-modal fusion is required to consider both vision and language information. Generally, VQA network follows two steps: 1) language understanding and image feature extraction; and 2) multi-modal fusion to generate the final representation. 
For language understanding, the most common way is to use recurrent neural network to generate the word embedding or sentence embedding \cite{ref-bua-vqa,ref-mcan}. Image feature extraction is greatly borrowed from the object detection network \cite{ref-bua-vqa}. 
 For the multi-modal fusion, several bilinear pooling methods were proposed for better integrated multi-modal features \cite{ben2017mutan}.  Recently, attention based multi-modal fusion is also proposed  \cite{ref-dynamic_fusion,ref-mcan,han2020finding}. 


\subsection{Self-Attention and Co-Attention based Methods}
Attention based approach has received much interest from the community.
Recently, Modular Co-Attention (MCA) structure was introduced to use self-attention of images and questions, as well as question-guided attention of images \cite{ref-mcan}. Later, works such as VideoBERT \cite{sun2019videobert} show that self-attention on both vision and language information can further improve the performance.
We share the similar idea of using attention for both vision and language. 
However, our work extends from the existing works by taking the answer representation back to perform the answer-checking process. As such, the multi-modal attention considers the current answer, question and image representation to generate new  representations. 
Therefore, our proposed multi-modal attention can unify the inter and intra attention of answer, question and image to improve the performance.


\section{Multi-modal Attention}

\subsection{Question-Guided Attention }
\label{qa-atn}

The attention function first computes the attention weights based on the softmax function and a predefined function (dot-product, sum, etc.), which takes a query and a set of key contents as input. Then the value contents are adaptively aggregated based on the attention weights. 
The question-guided attention is derived from \cite{ref-mcan} by assigning the question representation as the key and value, and the image representation as the query. More specifically, the question-guided attention output $I_{QG}$ is computed as follows: 
\begin{equation}
\begin{split}
\label {attention}
&Attn = softmax(\frac{Q_{image}K_{question}^T}{\sqrt{d_k}})V_{question} \\
&I_{QG} = LayerNorm(Attn +Q_{image})
\end{split}
\end{equation}
where attention input query $Q_{image}$ is set by the image feature  $ Y \in R^{B \times n_I \times d_c}$. The attention key $K_{question}$ and value $V_{question}$ are both set by the question feature  $X \in R^{B \times n_Q \times d_c}$, where $B$, $n_I$ and $n_Q$ represent the batch size, the number of tokens and bounding boxes respectively. 
Note that queries, keys and values are with dimensions of $d_q$, $d_k$ and $d_v$ respectively. In this paper, we set $d_q$=$d_v$=$d_k$=$d_c$ and use the common dimension $d_c$ to represent it. 

\begin{figure}[b]
	\centering 
	\includegraphics[width=0.48\textwidth]{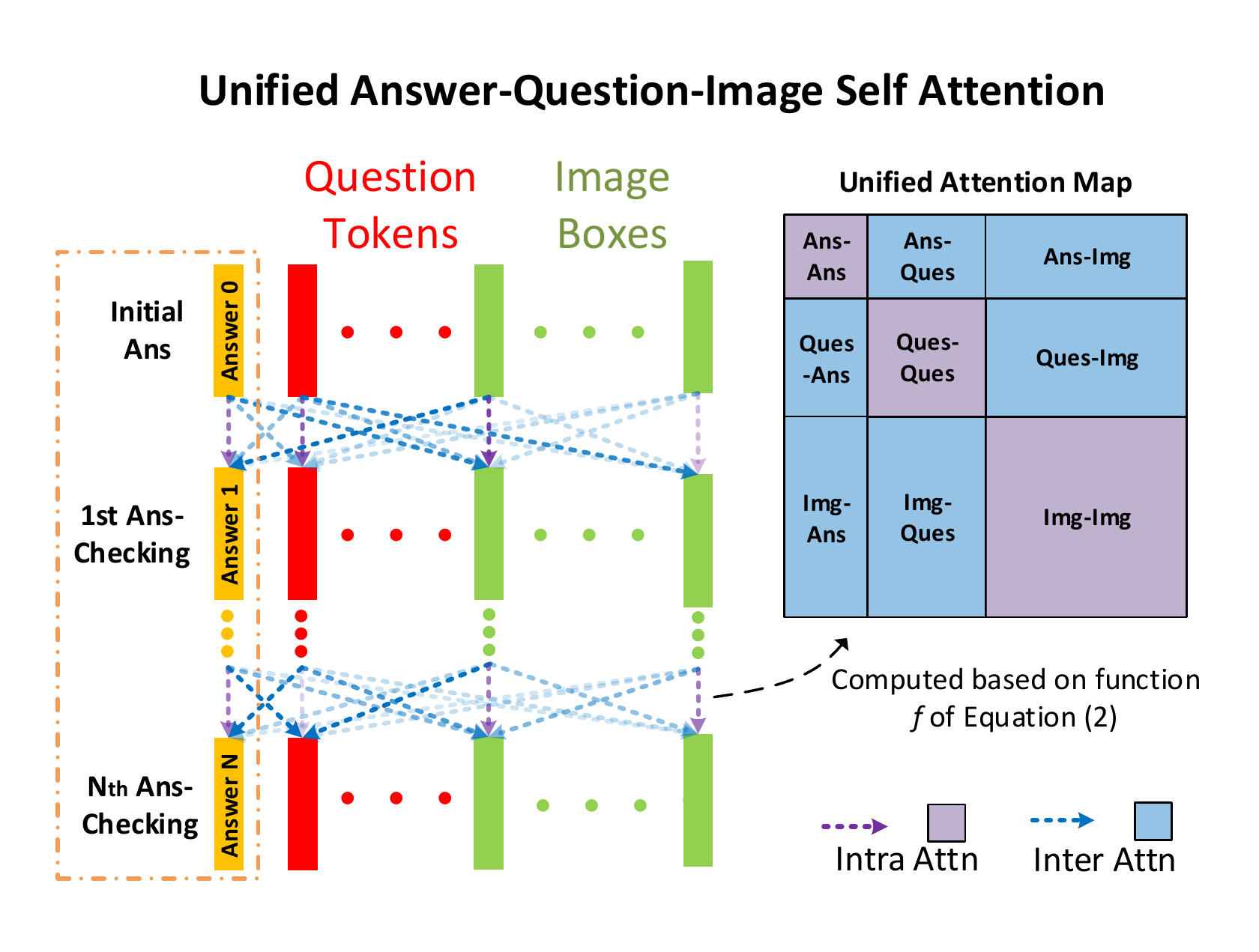}
	\caption{A unified answer-question-image self-attention with intra-and-inter modal attention. The answer-checking process is performed sequentially to generate the potential answer. }
	\label{fig:self_attention}
\end{figure}

\subsection{Proposed Unified Multi-modal Self-Attention}
In this paper, we propose an unified multi-modal self-attention. Such attention performs dynamic fusion of the answer representation $A$,  question representation $X$ and image representation $Y$. 
The answer representation $A$ is generated by averaging the concatenated question representation $X \in R^{B \times n_Q \times d_c}$ and image representation $Y \in R^{B \times n_I \times d_c}$. 
The fused features are represented by concatenating them as $Z$=$[A, X, Y] \in R^{B \times (n_Q+n_I+1) \times d_c}$. As such, the multi-modal self-attention is formulated as 

\begin{equation}
\label {IIM-att}
\begin{split}
& Attn(Z_Q, Z_K, Z_V)=f(Z_Q, Z_k) Z_V \\
&        = softmax(\frac{Z_QZ_K^T}{\sqrt{d_k}})Z_V
\end{split}        
\end{equation}
where $Z_Q$, $Z_K$ and $Z_V$ are independently linear projection of the fused feature $Z$. For example, $Z_Q = ZW_Q+B_Q$, where $W_Q$ and $B_Q$ are trainable weights and bias respectively. Such attention can be further viewed as a combination of inter-modal attention and intra-modal attention. 
The attention map is shown in Fig. \ref{fig:self_attention}, with intra-and-inter-modal attention of answer, question and image representation. This process can be clearer by using the submatrix $[Z_{Q_A} ,Z_{Q_X}, Z_{Q_Y}]$  to represent $Z_Q$ for attention as follow
\begin{equation}
\label {IIM-att2}
\begin{split}
Attn(Z_Q, Z_K, Z_V)= & f(\begin{bmatrix}
  Z_{Q_A} ,Z_{Q_X}, Z_{Q_Y}
\end{bmatrix},\\
& \begin{bmatrix}
  Z_{K_A}, Z_{K_X}, Z_{K_Y}
\end{bmatrix}) Z_V
\end{split}
\end{equation}
where $[Z_{Q_A} ,Z_{Q_X}, Z_{Q_Y}]$ represents the information from the answer, question and image. Similarly, $Z_K $= $[Z_{K_A}, Z_{K_X}, Z_{K_Y}]$.
$f$ represents the attention weight function, which is the scaled softmax in Equation (\ref{IIM-att}). As shown in Fig. \ref{fig:self_attention}, an unified answer-question-image self attention is developed based on Equation (\ref{IIM-att}) and (\ref{IIM-att2}). 
Such a unified attention is adopted sequentially in the answer-checking process to generate all the potential answers.

\begin{figure}[t]
	\centering 
	\includegraphics[width=0.43\textwidth]{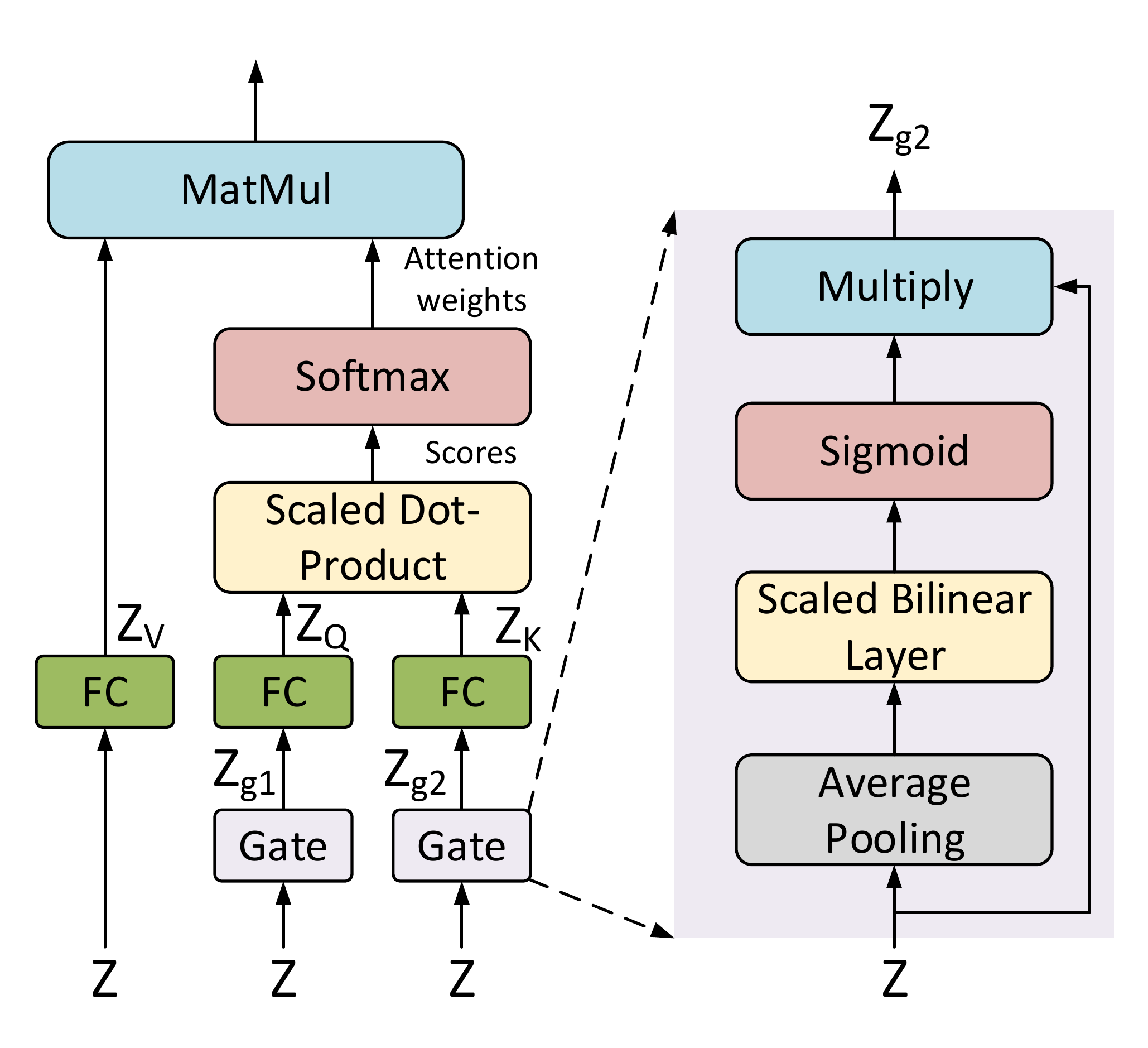}
	\caption{Self-attention with bilinear self-gated input}
	\label{fig:gate_design}
\end{figure}

\begin{figure*}[t]
	\centering 
	\includegraphics[width=0.9\textwidth]{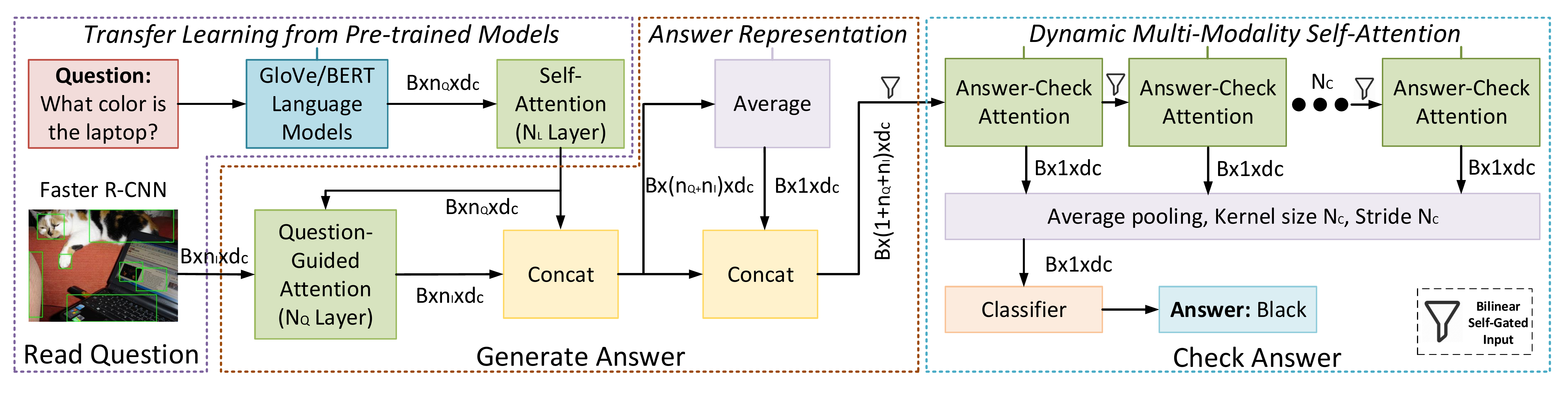}
	\caption{Overall architecture of proposed ARAC architecture with 3 subnetworks: Read question network with $N_L$ layers of question self-attention; Generate answer network with $N_Q$ layers of question-guided image attention and Check answer network with $N_c$ layers of multi-modal attention network. $B$ represents the batch size. Note that the answer representation is the average of question representation and image representation.}
	\label{fig:ml_arch}
\end{figure*}

\subsection{Proposed Bilinear Self-Gated Input}
Inspired by attention on attention \cite{huang2019attention} and bilinear attention networks \cite{kim2018bilinear}, we design a bilinear self-gated input gate to remove irrelevant features for the multi-modal attention module.

As shown in Fig. \ref{fig:gate_design}, the input is first fed into an average pooling. Then a bilinear projection is performed, followed by a sigmoid function to get the gated matrix $G$. 
The gated matrix is  broadcasted to perform elementwise multiplication with input $X$ to generate the output. The computation process can be summarized as follows. 
\begin{equation}
\label {bg_gate}
\begin{split}
& Z_p = avg\textnormal{-}pooling(Z, k, s), \  G = \sigma(g(Z_p, Z_p))  \\
& Z_g = Z \cdot G 
\end{split}
\end{equation}
where $k$ is the pooling kernel size and $s$ is the stride size resulting to a dimension reduction from $Z \in R^{B \times (n_Q+n_I+1) \times d_c}$ to $Z_p \in R^{B \times (n_Q+n_I+1) \times d_p}$. $Z_p$ is the average-pooled input features of input $Z$.
$g$ is a bilinear function defined as $g(x,y)$=$x^T My$, where
$M \in R^{d_p \times d_p \times 1}$ is a trainable weight. As such, the size of $G$ is further reduced to $Z \in R^{B \times (n_Q+n_I+1) \times 1}$ from $ R^{B \times (n_Q+n_I+1) \times d_p}$ to serve as a gate. The gated output matrix $Z_g \in R^{B \times (n_Q+n_I+1) \times d_c}$ is then computed by elementwise multiplication of input $Z$ and broadcasted input $G$. $Z_g$ is the gated input of $Z$. As shown in Fig. \ref{fig:ml_arch},
the bilinear self-gated input is added to the answer-check attention modules. 

\section{ARAC VQA Architecture}


\subsection{Image Representation}
In this paper, we propose an Attention based Read, Answer and Check (ARAC) architecture for visual question answering.
The input image features are extracted from bottom-up and top-down attention model \cite{ref-bua-vqa}. These features are obtained from a Faster R-CNN with ResNet101 \cite{ref-resent} trained on the object and attribute annotation from Visual Genome dataset \cite{ref-vg}. We adopt the same image feature extraction process as \cite{ref-bua-vqa} to generate potential image regions. The number of selected regions is in range $[10, 100]$ based on \cite{ref-bua-vqa} and we pad them to the fixed number of regions 100.
For each selected region, we perform a mean-pooled convolutional feature for that region such that the image feature dimension is set as 2048 \cite{ref-bua-vqa}.


\subsection{Read Question}

Question reading is designed based on intra-modal self-attention and further enhanced by pre-trained models. 
We set the maximum question length 14 for computational efficiency based on \cite{ref-bua-vqa}. The extra words are discarded and the short question is padded with zeros. We use transfer learning from pre-trained model to understand the question. 
For the GloVe pre-trained model \cite{ref-glove}, each word is represented by a 300-dimensional vector. 
The resulting sequence of word embedding will be $14 \times 300$. For the BERT \cite{ref-bert} pre-trained model, we follow exactly the same BERT pre-processing and use the open source pre-trained model \cite{huggingface}. The word vector dimension increases to 768 in this case. We do not use recurrent neural network as the previous works  \cite{ref-bua-vqa,ref-mcan} since it hurts the parallel computation and introduces large latency. 
The word embedding is first linearly converted to the common dimension $d_c$. Then an $N_L$ layers of language self-attention are stacked to compute the final word representation as shown in Fig. \ref{fig:ml_arch}.


\subsection{Generate Answer}
\label{gen_ans}
The generation process is performed by question-guided attention. 
We first adopt a fully connected (FC) layer to convert the image feature dimension $2048$ to the common dimension $d_c$. Then the image feature serves as attention input $Q$ and question representation is used as attention input $K$, $V$. The output of the question-guided attention is concatenated with language features as $Z_{QI}$=$[X,Y]$, where $Z_{QI} \in R^{B \times (n_Q+n_I) \times d_c}$. $n_Q$ and $n_I$ represent the number of image features (100) and number of words (14) respectively. 
The answer representation $A \in R^{B \times 1 \times d_c}$ is generated by computing the average of the concatenated representation $Z_{QI}$ along the second dimension.
Averaging word embedding is commonly used to represent the sentence \cite{joulin2016bag} . We borrow this idea to design a compact answer representation to represent the predicted answer. 
Finally, the answer representation is merged into the latent vector $Z \in R^{B\times (1+n_Q+n_I)\times d_c}$, composed of information from the answer, question and image.

\subsection{Check Answer}
\label{sec:checkans}
The answer-checking process is performed by the proposed multi-modal self-attention. 
The aforementioned latent vector $Z$ is sequentially fed into $N_C$ layers of answer-check attention modules as shown in Fig. \ref{fig:self_attention} and Fig.  \ref{fig:ml_arch}.
The answer representation $A$ is the first row of the latent vector $Z$. 
Each answer-check attention layer generates one answer representation according to the latent vector $Z$. Note that the latent vector $Z$ is updated layerwise through the answer-check attention modules.
As shown in Fig. \ref{fig:ml_arch}, the answer of each layer $A \in R^{B\times 1 \times d_c}$ is fed into an average pooling layer to make a conclusion.
The kernel size and stride of the average pooling layer is determined by the number of check attention layer $N_C$ to keep the output dimension the same as $d_c$. 
Note that average pooling is used instead of max pooling. This is because average pooling considers every answer thus can be regarded as majority voting, while max pooling tends to be overconfident which leads to overfitting.
This average-pooled answer will then go through a  linear classifier to infer the final answer. We adopt the binary cross entropy with logits as the cost function \cite{ref-bua-vqa}.

\begin{figure*}[t]
	\centering 
	\includegraphics[width=0.8\textwidth]{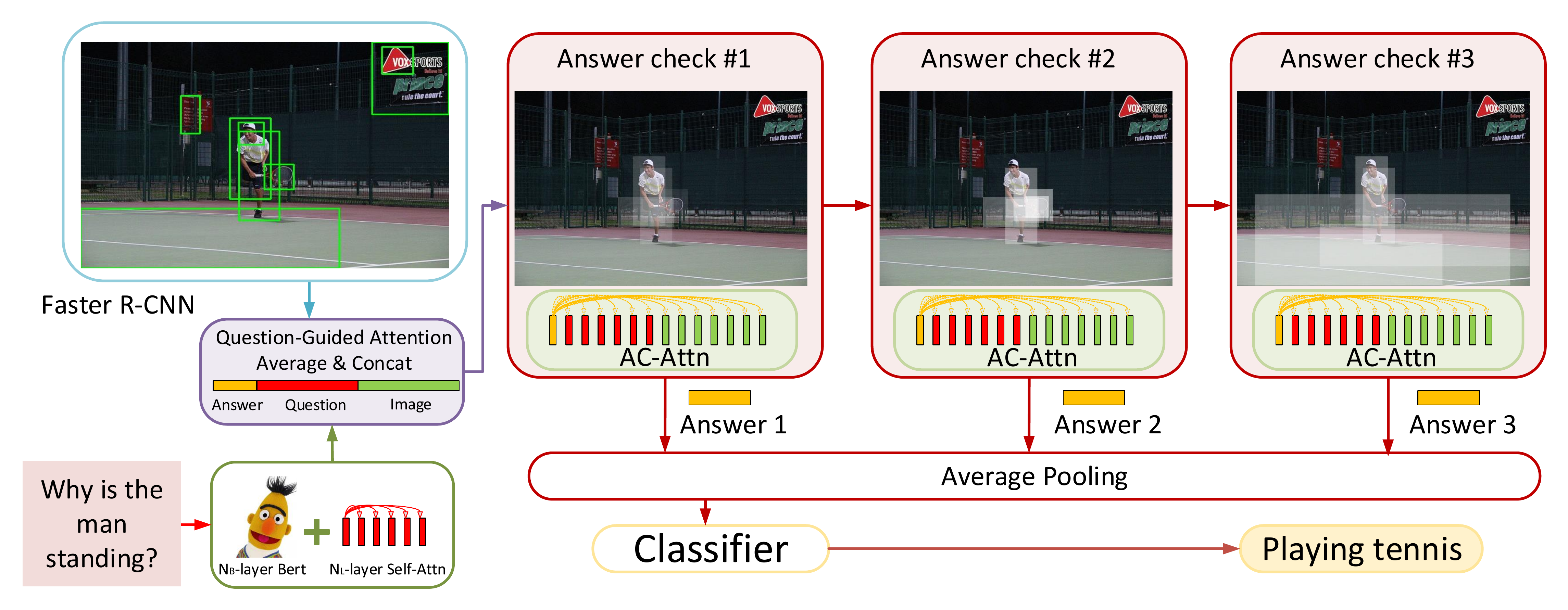}
	\caption{Visualized ARAC network working process for  VQA. In the answer-check attention (AC-attention) subnetwork, each answer representation specializes in some regions of the image. Then all the answer representation are average pooled to generate the summarized answer representation for the classifier.   }
	\label{fig:pooling_att}
\end{figure*}

Fig. \ref{fig:pooling_att} illustrates how these answers are generated and concluded with visualization of attention weights on the image. The question is ``Why is the man standing?", given an image of a sportsman playing tennis. In this case, we use three answer-check attention (AC-attention) layers as an example. 
In Fig. \ref{fig:pooling_att}, the transparency of the bounding box is based on the attention weights. The larger the weight is, the clearer the bounding box is. 
We observe that different answer representations focus on different regions of the images.

\section{Experiment}

\subsection{Dataset}
\label{data_intro}
In this paper, we use VQA-v2.0 \cite{ref-geogria_vqa} for model analysis. VQA-v2.0 dataset contains human annotated question-answer pairs for images from Microsoft COCO dataset \cite{ref-coco}. 
The answers of the test-standard set are only available on VQA-v2.0 test server. 
We follow \cite{ref-bua-vqa}, and restrict the set of candidate answers to correct answers in the training set that appears more than 8 times, leading to $d_{ca}$=3129 candidate answers.  

We also evaluate our model using the VQA-CP v2 dataset \cite{vqa-cp}, which is designed with a different distribution of train and test split from VQA v2.0 dataset for overcoming the language priors.
We further evaluate our architecture on Vizwiz dataset \cite{vizwiz}.  VizWiz dataset was collected by the visually-impaired people using mobile phones.
The Vizwiz dataset differs from the conventional dataset from two perspectives: 1) the image could be blur; 2) the question could be unrelated to the image. This is to simulate the question asked and images taken by blind people. The vizwiz dataset is divided into training, validation, and test sets of 20,000, 3,173, and 8,000
visual questions, respectively. 

\subsection{Implementation Details}
To simplify the hyperparameter choosing process, we set the number of blocks to be equal , $N_L$=$N_Q$=$N_C$. We convert the image features from Faster R-CNN with dimension 2048 to the common dimension $d_c$=$512$. For language models, we convert the GloVe model word vector 300 dimensions or BERT model 768 dimensions to $d_c$=$512$. The default language model will be the GloVe model in the following experiments. 
For the attention configuration, we use 8 attention heads for all the experiments. In the bilinear self-gated input network, the pooling kernel size and stride are set the same as the number of attention heads, 8.
For the fully connected layer in the attention block, we first convert the dimension to $2*d_c$ and then convert it back to $d_c$. 
Our architecture is implemented in PyTorch \cite{pytorch} with Adamax optimizer \cite{kingma2014adam}. 
The basic learning rate is $lr$=$0.002$. The first 4 epochs are warm-up, which follows $lr/(5-epoch)$. We keep the basic learning rate until epoch 20 and then decrease it to $lr = 0.1*lr$.  The learning rate of BERT parameters is always set as $0.01*lr$.

\begin{figure*}[!t]
	\centering 
	\includegraphics[width=0.8 \textwidth]{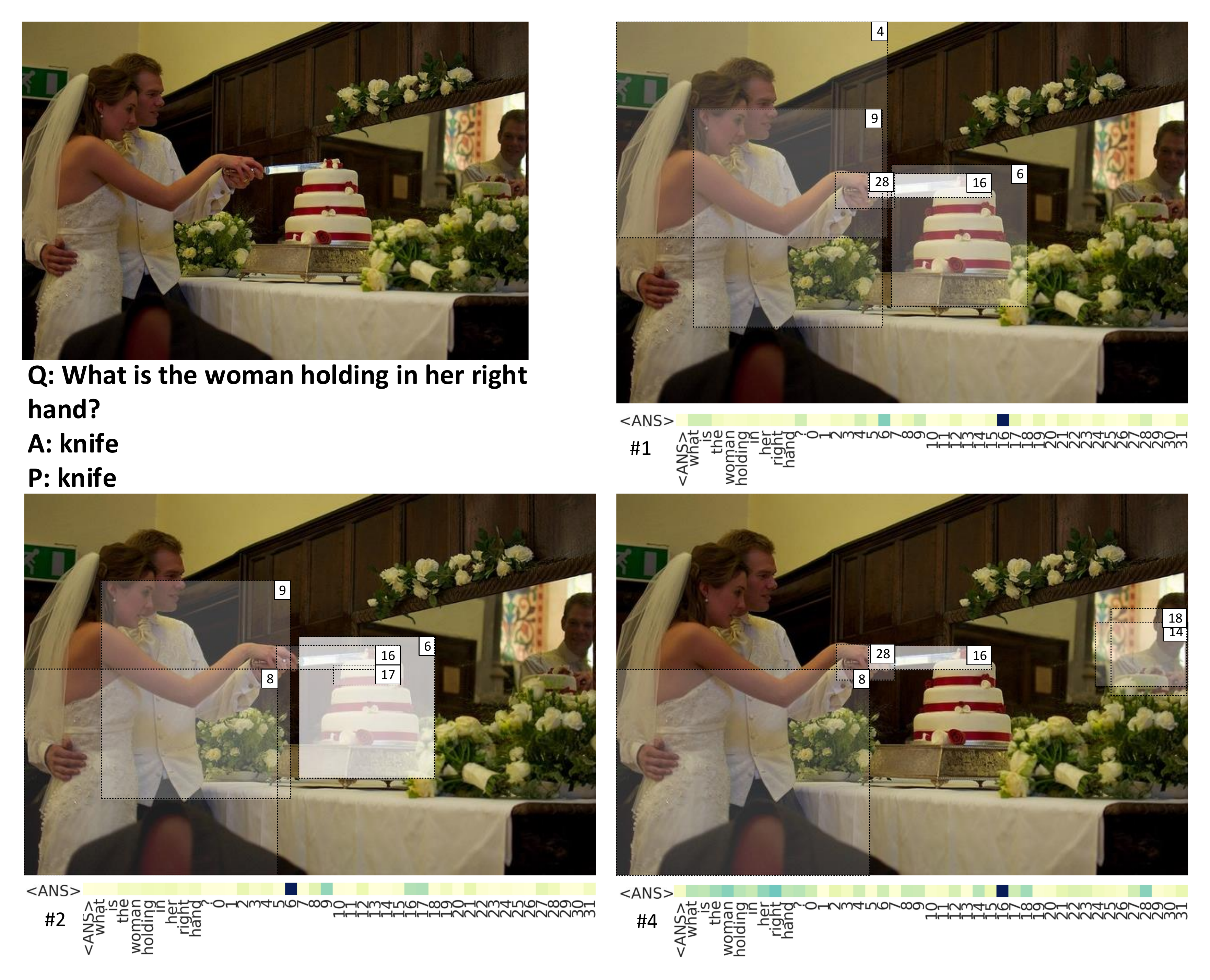}
	\caption{Visualization of answer-check attention weights for ARAC-5 architecture. We take the layer 1, 2 and 4 AC-Attention weights and visualize them through the bounding box. Clearer bounding box represents larger attention weight. 
	The heatmap below each image shows the answer attention map to the current answer, question and image representation. In the heatmap, a darker color represents a larger value. The bounding boxes with top-5 largest attention weight are drawn on the image. More visualizations are in the supplementary materials. Best viewed in 200\%;}
	\label{fig:att4}
\end{figure*}

\subsection{Ablation Studies}
\label{aba_discuss}

\begin{table}[t]
\small
\caption{Ablation studies of our proposed ARAC on VQA
v2.0 validation dataset. Best performance is highlighted in bold. }
\label{ablation}
\begin{center}
\begin{tabular}{ |l c c | } 
\hline
Component & Setting & Acc (\%) \\
\hline
   {Bottom-up \cite{ref-bua-vqa}}  & Bottom-up &  63.37\\
\hline
\multirow{1}{10em}{Bilinear Attention  \cite{kim2018bilinear}} 
& BAN-12 & 66.04 \\
\hline
\multirow{2}{9em}{DFAF \cite{ref-dynamic_fusion}} 
& DFAF-8 & 66.66 \\ 
& DFAF-8-BERT & \textbf{67.23} \\ 
\hline
MLI-BERT  \cite{gao2019multi} & Latent Interaction   & 67.83 \\
\hline
Default  &  ARAC-4  & 66.89 \\
\hline
\multirow{4}{9em}{Number of Stacked Blocks} & ARAC-3 & 66.76 \\
& ARAC-4 & \textbf{66.89} \\ 
& ARAC-5 & 66.79 \\ 
& ARAC-6 & 66.88 \\ 
\hline
\multirow{3}{9em}{Number of Hidden Nodes} & 256 & 66.18 \\
& 512 & \textbf{66.89 }\\ 
& 768 & 66.84 \\ 

\hline
\multirow{2}{9em}{Attention Gating} & No gating & 66.82 \\
& Bilinear Self-gate & \textbf{66.89} \\ 

\hline
\multirow{5}{9em}{Number of AC-attention Layers} & 0 layer & 60.24 \\
& 1 layer & 66.65  \\ 
& 4 layers & \textbf{66.89} \\
& 6 layers & 66.74 \\
\hline
\multirow{2}{9em}{Answer Token} & With Answer Token & \textbf{66.89} \\
& Without Answer Token & 66.74  \\  \hline
\multirow{4}{9em}{Positional Encoding}
& Image & 66.57 \\
& Question  & 66.71  \\ 
& Question and Image & 66.40\\
\hline
\multirow{3}{9em}{BERT Layer 1} & ARAC-3& 67.47 \\
& ARAC-4 & \textbf{67.48} \\
& ARAC-5 & 67.45 \\ 
 \hline
\multirow{3}{9em}{BERT Layer 6}  & ARAC-3 & 67.99 \\
& ARAC-4 & 68.00 \\ 
& ARAC-5 & \textbf{68.14} \\ 
 \hline
\multirow{2}{9em}{BERT Layer 12} 
& ARAC-3 & 67.75 \\ 
& ARAC-4 & 67.74 \\ 
& ARAC-5 & \textbf{67.95} \\ 
\hline
Proposed \textbf{Tiny Model} & ARAC-4-GloVe & 66.89 \\
\hline
Proposed \textbf{Base Model} & ARAC-4-BERT-1 & 67.48 \\
\hline
Proposed \textbf{Large Mode}l & ARAC-5-BERT-6 & \textbf{68.14}\\

\hline
\end{tabular}
\end{center}

\end{table}

We perform extensive ablation studies on the VQA-v2.0 validation dataset \cite{ref-geogria_vqa} using the train split dataset for training only. 
The results are summarized in Table \ref{ablation}.
The default setting of the ablation study is ARAC-4, which represents 4 blocks of language attention, question-guided attention and check attention respectively ($N_L$=$N_Q$=$N_C$=$4$) using GloVe word embedding. The common dimension $d_c$ (number of hidden nodes) is 512.


\textbf{Effects of Network Configuration}
The ARAC network is not very sensitive to the network depth. The performance slightly varies as the depth increases. 
On the other hand, the ARAC network is more sensitive to the width. Reducing the common dimension by half to $d_c$=256, the performance drops around 0.7\%. Increasing the width will also not benefit the performance. This shows common dimension $d_c$=512 is a local optimal hyperparameter.  The proposed bilinear input gate improves the performance from 66.82\% to 66.89\% with a negligible model size increase. The parameter number of bilinear weight is only $d_c^2/64$ since the dimension is reduced by average pooling operation with kernel size 8 and stride 8.  

\textbf{Effects of Positional Encoding}
We also investigate the learnt positional encoding for both question and image. The question positional encoding is based on \cite{ref-transformer} and 2D image positional encoding is performed based on  \cite{carion2020end}, where each object coordinate ($x_{min}$, $y_{min}$, $x_{max}$, $y_{max}$) is mapped to a $d_c$ (512) dimensional feature. The 4 coordinate embeddings are then concatenated to represent the final positional embedding with dimension 2048, same as the image feature dimension. 
Detailed implementation could be found at \cite{carion2020end}.
We find that positional embedding leads to a small accuracy drop from 66.89\% to 66.40\%. 
This indicates that the long range dependency in the question may not be as important as other NLP tasks  such as neural machine translation.
The image positional information may be under-trained due to insufficient training data relating to image positional information since many questions may not require positional information to answer. We leave the investigation for our future works. 

\textbf{Effects of Answer Checking.}
Since ``Answer Checking Module" is the most important part in our model, 
we have thoroughly investigated its effectiveness.
Firstly we investigate the depth benefits by varying the AC-attention layers from 0 to 6 as shown in Table \ref{ablation}. When there is no AC-attention layer, the initial answer representation is directly fed into the classifier, resulting an accuracy 60.24\% on validation dataset.
However, with only 1 layer of AC attention, the performance is significantly improved from 60.24\% to 66.65\%, which verifies the effectiveness of the AC attention. We further adopt 4 and 6 layers of ``answer checking" modules, which can achieve 66.89\% and 66.73\%  accuracy respectively. This shows deeper AC attention layers can improve the performance until the benefits degrade after 4 layers.
We have also investigated the needs of answer representation. As discussed in Section \ref{gen_ans}, the input for the answer-checking module is $[A, X, Y]$, where $A$, $X$, $Y$ represent the answer, question and image representation respectively.  If we feed concatenated image and question $[X, Y]$ without answer token directly to the ``answer checking" module, the accuracy drops to 66.74\%. This indicates an answer representation helps to extract the related language and image information for the final classifier.


\textbf{Effects of Transferring BERT.}
We investigate the transfer learning of BERT \cite{ref-bert} to replace GloVe\cite {ref-glove}. 
ARAC-4-BERT-1 and ARAC-4-BERT-6 refer to 1 BERT-base layer and 6 BERT-base layers respectively.
Our results show that even with only 1 layer of BERT transferred (ARAC-4-BERT-1), there is significant performance improvement from 66.89\% to 67.48\% as shown in Table \ref{ablation}.
However, not all the BERT layers are needed.  For the ARAC-3 case, when all the BERT layers (12 layers) are transferred, the accuracy is around 67.75\%. This is only slightly better than transferring 1 BERT layer. However, transferring 6 BERT layer (ARAC-4-BERT-6)  achieves the best result around 67.99\% accuracy.
This indicates that the earlier BERT layers (from Layer 1 to Layer 6) is more for the general word embedding. 
Therefore, our \textbf{Large Model} is set based on 6 BERT layers transferred. 



\subsection{Comparison with the State-of-the-art Works}
\label{subsec:cmp}
We compare the model size with consideration of accuracy in Table \ref{model_size}. Furthermore, the accuracy comparison with the state-of-the-art works on various datasets is summarized in Table \ref{cmp_sota} and Table \ref{vizwiz-sota}.
In Table \ref{model_size}, size ratio is calculated based on our base mode (ARAC-4-BERT-1). Clearly,  our tiny model (ARAC-4-GloVe) is the smallest among the listed models while achieves better performance than BAN-4 and DFAF. Compared to  MUAN-768 \cite{yu2019multimodal},
our model (ARAC-4-BERT-1) not only achieves performance improvement (0.2\% accuracy increase) but also uses 28\% fewer parameters. 
Note that since the vocabulary of VQA2.0 dataset is small, only a part of GloVe embedding is used for word embedding resulting in $6.84M$ number of parameters. For BERT, the word embedding takes $15.63M$ number of parameters. 
We also observe that our \textbf{large model} achieves better accuracy as the size grows compared to \textbf{base model}.
It is worthy to mention that only 0.02\% accuracy improvement is achieved when MUAN-768  size is increased twice to MUAN-1024. 



In Table \ref{cmp_sota}, our \textbf{Large Model} ARAC-5-BERT-6 (using 6 BERT layers) achieves a better accuracy 71.57\% with fewer number of parameters compared to VilBERT \cite{vibert} 70.9\%. VilBERT \cite{vibert} model size is 267M parameters, which is significantly larger than our \textbf{Large Model} 101.0M parameters.
Fig. \ref{fig:intro1} shows a clear model performance (model size and accuracy) comparison with existing works. Through the  \textbf{Tiny}, \textbf{Base} and \textbf{Large} model, our model shows a larger slope of accuracy over model size compared to the DFAF model. 
This indicates that our architecture has a better utilization of parameters for better performance than existing works. Moreover, 
\textbf{Large Model} achieves the state-of-the-art single-model performance (71.57\%) on VQA-v2.0 test-std dataset. 
For the fair comparison, we do not compare with models that use significant amount of additional data for pre-training such as LXMERT \cite{lxmert} and VIBERT \cite{vibert}. For example, LXMBERT used 3 pre-train tasks: 1) masked language modeling with 100M words trained as BERT \cite{ref-bert}; 2) masked object prediction with 6.5M additional objects  and 3) visual-question answering task with 9.18M image-sentence pairs and 180K distinct images. 
These models use much more data compared to ours using VQA 2.0 with 650K image-question pairs and 120K images.


In Table \ref{vqacp}, we show the performance on the VQA-CP v2 dataset \cite{vqa-cp}.  As shown in Table \ref{vqacp}, we achieve a nearly state-of-the-art accuracy 43.4\% compared to the existing works. Rubi \cite{cadene2019rubi} learning method can further improve the performance for this dataset.
In Table \ref{vizwiz-sota}, we also compare our model with existing works on Vizwiz \cite{vizwiz} dataset. Please note that the Vizwiz is more difficult  than VQA-v2.0 since it includes an additional question type (unanswerable) and many blur images. 
Our model works well on Yes/No type questions. However, for the unanswerable type of questions, it is less impressive compared to BAN model \cite{ban-vizwiz}. 
This may be due to the answer-checking modules, which try to inference some answers for unanswerable type of questions.
Our model achieves an overall accuracy of 53.4\%, which ourperforms the existing methods.
Note that we do not compare with Pythia \cite{textvqa} since it uses additional training datasets such as Visual-Genome \cite{ref-vg} and  Visual-Dialog \cite{visual_dialogue}.

\begin{table}[!t]
\small
\caption{ 
Number of model parameters comparison including word embedding on VQA-v2.0 validation dataset }
\label{model_size}
\begin{center}
\begin{tabular}{ |l c  c c | } 
\hline
Model Name & Model Size & Size Ratio & Acc (\%) \\
\hline
BAN-4  \cite{kim2018bilinear}  & 44.8M & 0.76 & 65.81\\
\hline
MCAN-6\cite{ref-mcan}  & 56M  & 0.95 & 67.20\\
\hline
MUAN-768 \cite{yu2019multimodal}   & 83M & 1.40 & 67.28\\
\hline
MUAN-1024 \cite{yu2019multimodal}   & 141.6 & 2.39 & 67.30\\
\hline
{DFAF \cite{ref-dynamic_fusion}} & 63.2M & 1.07 & 66.66\\
\hline
{DFAF-BERT \cite{ref-dynamic_fusion}} & 173.2M & 2.93 & -\\
\hline
{MLI-BERT  \cite{gao2019multi}}& 120M & 2.03 & 67.83\\
\hline
\textbf{ARAC-4-GloVe} & 34.4M & 0.58 & 66.89 \\
\hline
\textbf{ARAC-4-BERT-1}  & 59.2M & 1 & 67.48 \\
\hline
\textbf{ARAC-5-BERT-6}  & 101.0M & 1.71 & \textbf{68.14}\\
\hline
\end{tabular}
\end{center}

\end{table}

\begin{table}[t]
\small
\caption{State-of-the-art accuracy comparison on a \textbf{single-model} for the test-dev and test-standard splits. The results are collected from the VQA-v2.0 competition server. 
The model is trained on VQA v2.0 training dataset, validation dataset and visual genome dataset.
}

\label{cmp_sota}
\begin{center}
\begin{tabular}{ |l |c c c c |c|  } 

\hline
   \multirow{2}{6em}{Model}  &  \multicolumn{4}{c|}{test-dev} & test-std \\ 
   &  Y/N & No. &Other & All   &All \\
\hline
{BUA\cite{ref-bua-vqa}}  &81.8 & 44.2 & 56.1 &65.32 &65.67\\

\hline
{BAN \cite{kim2018bilinear}}  &85.3 & 50.9 & 60.3 & 69.52 & -- \\

\hline
{BAN-C \cite{kim2018bilinear}}  &85.4 & 54.0 & 60.5 & 70.04&70.35 \\

\hline
{DFAF \cite{ref-dynamic_fusion}}  &86.1 & 53.3 & 60.5 &70.22 &70.34 \\

 \hline
{DFAF-BERT  \cite{ref-dynamic_fusion}}  &86.7 & 52.9 & 61.0 &70.59& 70.81 \\

 \hline
 {MCAN  \cite{ref-mcan}}  &86.8 & 53.3 & 60.7 &70.63 &70.90 \\

 \hline
 {MUAN \cite{yu2019multimodal}}  &86.8 & \textbf{54.4} & 60.9 &70.82 &71.10 \\
 \hline
 {MLI  \cite{gao2019multi}}  &86.0 & 52.9 & 60.4 &71.28 &70.28 \\
 \hline
 {MLI-BERT  \cite{gao2019multi}}  &87.1 & 53.4 & 60.5 &71.09 &71.27 \\
 \hline
 {Unified-VLP  \cite{UVLP}}  &87.4 & 52.1 & 60.5 &70.6 &70.7 \\
  \hline
 {Visual-BERT \cite{li2019visualbert}}  &-- & -- & -- &70.8 &71.0 \\
\hline
 {VilBERT \cite{vilbert}}  &-- & -- & -- &70.6 &70.9 \\
\hline
 {QBN \cite{shi2020multi}}  &87.1 & 52.93 & 60.8 &70.8 &71.0 \\
\hline
\textbf{ARAC-4-GloVe} &85.9 & 52.5 & 60.5 & 70.06 & 70.40 \\
\hline
\textbf{ARAC-4-BERT-1}  &86.8 & 53.0 & 61.2 &70.81   & 71.03 \\
\hline
\textbf{ARAC-5-BERT-6} &\textbf{87.4} & 54.1 & \textbf{61.6} &\textbf{71.34 }& \textbf{71.57} \\

\hline
\end{tabular}
\end{center}
\end{table}

\begin{table}[hbt]
	\small
	\centering
	\caption{\label{vqacp}
			State-of-the-art accuracy(\%) comparison on a \textbf{single-model} for the  VQA-CP v2 \cite{vqa-cp} test split. BUA result is from \cite{huang2020aligned}. BAN and MCAN results are from  \cite{kervadec2020estimating}.
		}

	\begin{tabular}{|c|c|c|c|c|c|}
		\hline
		Methods   & \begin{tabular}[c]{@{}c@{}}BUA \\ \cite{ref-bua-vqa} \end{tabular} & \begin{tabular}[c]{@{}c@{}}BAN\\  \cite{kim2018bilinear}
		\end{tabular} & \begin{tabular}[c]{@{}c@{}}DC-GCN\\ \cite{huang2020aligned} \end{tabular} & \begin{tabular}[c]{@{}c@{}}MCAN\\ \cite{yu2019multimodal} \end{tabular} & \begin{tabular}[c]{@{}c@{}}\textbf{ARAC-4} \\ \textbf{(Ours) }\end{tabular} \\ \hline
		Test Acc. & 39.74                                                 & 38.01                                                & 41.47                                                   & 42.5                                                  & \textbf{43.4}   \\ \hline
	\end{tabular}
\end{table}

\begin{table}[t]
\small
\caption{State-of-the-art accuracy(\%) comparison on a \textbf{single-model} for the  Vizwiz test-standard split. The training dataset is based on the \textbf{Vizwiz train and validation split only}. }

\label{vizwiz-sota}
\begin{center}
\begin{tabular}{ |l |c c c c c|  } 
\hline
Model    & Yes/No & No. & Unans &Other & All \\
\hline
{Q+I+BUA \cite{vizwiz}}  &58.2 & 7.1 & 6 &14.3 & 13.4\\
\hline
{VizWiz\cite{vizwiz}}  &59.6 & 21.0 & 80.5 & 27.3 &46.9\\
\hline
{BAN \cite{ban-vizwiz}}  &68.1 & 17.9 & \textbf{85.3} & 31.5 & 51.6 \\
\hline
{\textbf{ARAC-5-BERT-6}}  &\textbf{73.4} & \textbf{26.7} & 82.6 & \textbf{35.4} & \textbf{53.4} \\
\hline

\end{tabular}
\end{center}
\end{table}

\subsection{Visualization}
We visualize the AC-attention weights in Fig. \ref{fig:att4}. 
The heatmap under each image represents the answer attention to the current answer, question and image representation for answer-checking layer 1, 2 and 4. The answer attention is extracted from the first row of the unified attention map as discussed in Fig. \ref{fig:self_attention}. 
As discussed in Section \ref{sec:checkans},
different answer representations focus on different regions of the image to answer the question.
As shown in Fig. \ref{fig:att4}, the answer representation from AC-attention Layer 1 pays heavy attention to the knife.  In AC-attention Layer 2, the answer representation pays more attention to the cake, which is also near the right hand of the woman. In AC-attention Layer 4, the answer representation again focuses on the knife. Note that bounding box 28 is the woman's right hand. 
This visualization matches the human intuition for answer checking.

\section{Conclusion}
 
 In this paper, we propose a fully attention based VQA architecture.
 Moreover, an answer-checking module is proposed to perform a unified attention on the jointly answer, question and image representation to update the answer.
 This mimics the human answer checking process to consider the answer in the context. With answer-checking modules and transferred BERT layers, our model achieves the state-of-the-art accuracy 71.57\%  with a smaller model size compared to the current sate-of-the-art models.

\begin{small}
{
	\bibliographystyle{IEEEtran}
	\bibliography{egbib_short}

\begin{thebibliography}{10}
\providecommand{\url}[1]{#1}
\csname url@samestyle\endcsname
\providecommand{\newblock}{\relax}
\providecommand{\bibinfo}[2]{#2}
\providecommand{\BIBentrySTDinterwordspacing}{\spaceskip=0pt\relax}
\providecommand{\BIBentryALTinterwordstretchfactor}{4}
\providecommand{\BIBentryALTinterwordspacing}{\spaceskip=\fontdimen2\font plus
\BIBentryALTinterwordstretchfactor\fontdimen3\font minus
  \fontdimen4\font\relax}
\providecommand{\BIBforeignlanguage}[2]{{%
\expandafter\ifx\csname l@#1\endcsname\relax
\typeout{** WARNING: IEEEtran.bst: No hyphenation pattern has been}%
\typeout{** loaded for the language `#1'. Using the pattern for}%
\typeout{** the default language instead.}%
\else
\language=\csname l@#1\endcsname
\fi
#2}}
\providecommand{\BIBdecl}{\relax}
\BIBdecl

\bibitem{ref-transformer}
A.~Vaswani, N.~Shazeer, N.~Parmar, J.~Uszkoreit, L.~Jones, A.~N. Gomez,
  {\L}.~Kaiser, and I.~Polosukhin, ``Attention is all you need,'' in
  \emph{NIPS}, 2017, pp. 5998--6008.

\bibitem{ref-bert}
J.~Devlin, M.-W. Chang, K.~Lee, and K.~Toutanova, ``{BERT}: Pre-training of
  deep bidirectional transformers for language understanding,'' \emph{arXiv
  preprint arXiv:1810.04805}, 2018.

\bibitem{ref-dynamic_fusion}
P.~Gao, Z.~Jiang, H.~You, P.~Lu, S.~C. Hoi, X.~Wang, and H.~Li, ``Dynamic
  fusion with intra-and inter-modality attention flow for visual question
  answering,'' in \emph{CVPR}, 2019, pp. 6639--6648.

\bibitem{li2019visualbert}
L.~H. Li, M.~Yatskar, D.~Yin, C.-J. Hsieh, and K.-W. Chang, ``Visualbert: A
  simple and performant baseline for vision and language,'' \emph{arXiv
  preprint arXiv:1908.03557}, 2019.

\bibitem{vilbert}
J.~Lu, D.~Batra, D.~Parikh, and S.~Lee, ``Vilbert: Pretraining task-agnostic
  visiolinguistic representations for vision-and-language tasks,'' in
  \emph{NIPS}, 2019, pp. 13--23.

\bibitem{gao2019multi}
P.~Gao, H.~You, Z.~Zhang, X.~Wang, and H.~Li, ``Multi-modality latent
  interaction network for visual question answering,'' in \emph{ICCV}, 2019,
  pp. 5825--5835.

\bibitem{ref-bua-vqa}
P.~Anderson, X.~He, C.~Buehler, D.~Teney, M.~Johnson, S.~Gould, and L.~Zhang,
  ``Bottom-up and top-down attention for image captioning and visual question
  answering,'' in \emph{CVPR}, 2018, pp. 6077--6086.

\bibitem{ref-mcan}
Z.~Yu, J.~Yu, Y.~Cui, D.~Tao, and Q.~Tian, ``Deep modular co-attention networks
  for visual question answering,'' in \emph{CVPR}, 2019, pp. 6281--6290.

\bibitem{ben2017mutan}
H.~Ben-Younes, R.~Cadene, M.~Cord, and N.~Thome, ``{MUTAN}: Multimodal tucker
  fusion for visual question answering,'' in \emph{ICCV}, 2017, pp. 2612--2620.

\bibitem{han2020finding}
W.~Han, H.~Huang, and T.~Han, ``Finding the evidence: Localization-aware answer
  prediction for text visual question answering,'' \emph{arXiv preprint
  arXiv:2010.02582}, 2020.

\bibitem{sun2019videobert}
C.~Sun, A.~Myers, C.~Vondrick, K.~Murphy, and C.~Schmid, ``{VideoBERT}: A joint
  model for video and language representation learning,'' \emph{arXiv preprint
  arXiv:1904.01766}, 2019.

\bibitem{huang2019attention}
L.~Huang, W.~Wang, J.~Chen, and X.-Y. Wei, ``Attention on attention for image
  captioning,'' \emph{arXiv preprint arXiv:1908.06954}, 2019.

\bibitem{kim2018bilinear}
J.-H. Kim, J.~Jun, and B.-T. Zhang, ``Bilinear attention networks,'' in
  \emph{NIPS}, 2018, pp. 1564--1574.

\bibitem{ref-resent}
S.~Ren, K.~He, R.~Girshick, and J.~Sun, ``{Faster R-CNN}: Towards real-time
  object detection with region proposal networks,'' in \emph{NIPS}, 2015, pp.
  91--99.

\bibitem{ref-vg}
R.~Krishna, Y.~Zhu, O.~Groth, J.~Johnson, K.~Hata, J.~Kravitz, S.~Chen,
  Y.~Kalantidis, L.-J. Li, D.~A. Shamma \emph{et~al.}, ``{Visual Genome}:
  Connecting language and vision using crowdsourced dense image annotations,''
  \emph{IJCV}, vol. 123, no.~1, pp. 32--73, 2017.

\bibitem{ref-glove}
J.~Pennington, R.~Socher, and C.~Manning, ``{GloVe}: Global vectors for word
  representation,'' in \emph{EMNLP}, 2014, pp. 1532--1543.

\bibitem{huggingface}
H.~Face, ``pytorch-transformers,''
  \url{https://github.com/huggingface/pytorch-transformers}, 2019.

\bibitem{joulin2016bag}
A.~Joulin, E.~Grave, P.~Bojanowski, and T.~Mikolov, ``Bag of tricks for
  efficient text classification,'' \emph{arXiv preprint arXiv:1607.01759},
  2016.

\bibitem{ref-geogria_vqa}
Y.~Goyal, T.~Khot, D.~Summers-Stay, D.~Batra, and D.~Parikh, ``Making the {V}
  in {VQA} matter: Elevating the role of image understanding in visual question
  answering,'' in \emph{CVPR}, 2017, pp. 6904--6913.

\bibitem{ref-coco}
T.-Y. Lin, M.~Maire, S.~Belongie, J.~Hays, P.~Perona, D.~Ramanan,
  P.~Doll{\'a}r, and C.~L. Zitnick, ``Microsoft {COCO}: Common objects in
  context,'' in \emph{ECCV}.\hskip 1em plus 0.5em minus 0.4em\relax Springer,
  2014, pp. 740--755.

\bibitem{vqa-cp}
A.~Agrawal, D.~Batra, D.~Parikh, and A.~Kembhavi, ``Don't just assume; look and
  answer: Overcoming priors for visual question answering,'' in \emph{CVPR},
  2018.

\bibitem{vizwiz}
D.~Gurari, Q.~Li, A.~J. Stangl, A.~Guo, C.~Lin, K.~Grauman, J.~Luo, and J.~P.
  Bigham, ``Vizwiz grand challenge: Answering visual questions from blind
  people,'' in \emph{CVPR}, 2018, pp. 3608--3617.

\bibitem{pytorch}
A.~Paszke, S.~Gross, S.~Chintala, G.~Chanan, E.~Yang, Z.~DeVito, Z.~Lin,
  A.~Desmaison, L.~Antiga, and A.~Lerer, ``Automatic differentiation in
  pytorch,'' 2017.

\bibitem{kingma2014adam}
D.~P. Kingma and J.~Ba, ``Adam: A method for stochastic optimization,''
  \emph{arXiv preprint arXiv:1412.6980}, 2014.

\bibitem{carion2020end}
N.~Carion, F.~Massa, G.~Synnaeve, N.~Usunier, A.~Kirillov, and S.~Zagoruyko,
  ``End-to-end object detection with transformers,'' \emph{arXiv preprint
  arXiv:2005.12872}, 2020.

\bibitem{yu2019multimodal}
Z.~Yu, Y.~Cui, J.~Yu, D.~Tao, and Q.~Tian, ``Multimodal unified attention
  networks for vision-and-language interactions,'' \emph{arXiv preprint
  arXiv:1908.04107}, 2019.

\bibitem{vibert}
W.~Su, X.~Zhu, Y.~Cao, B.~Li, L.~Lu, F.~Wei, and J.~Dai, ``Vl-bert:
  Pre-training of generic visual-linguistic representations,'' 2020.

\bibitem{lxmert}
H.~Tan and M.~Bansal, ``Lxmert: Learning cross-modality encoder representations
  from transformers,'' in \emph{EMNLP-IJCNLP}, 2019.

\bibitem{cadene2019rubi}
R.~Cadene, C.~Dancette, M.~Cord, D.~Parikh \emph{et~al.}, ``Rubi: Reducing
  unimodal biases for visual question answering,'' in \emph{NIPS}, 2019, pp.
  841--852.

\bibitem{ban-vizwiz}
J.-H. Kim, Y.~Choi, S.~Hong, J.~Jun, and B.-T. Zhang, ``Bilinear attention
  networks for vizwiz challenge.'' in \emph{ECCV Workshop}, 2018.

\bibitem{textvqa}
A.~Singh, V.~Natarajan, M.~Shah, Y.~Jiang, X.~Chen, D.~Batra, D.~Parikh, and
  M.~Rohrbach, ``Towards vqa models that can read,'' in \emph{CVPR}, 2019, pp.
  8317--8326.

\bibitem{visual_dialogue}
A.~Das, S.~Kottur, K.~Gupta, A.~Singh, D.~Yadav, J.~M. Moura, D.~Parikh, and
  D.~Batra, ``Visual dialog,'' in \emph{CVPR}, 2017, pp. 326--335.

\bibitem{UVLP}
L.~Zhou, H.~Palangi, L.~Zhang, H.~Hu, J.~J. Corso, and J.~Gao, ``Unified
  vision-language pre-training for image captioning and vqa,'' in \emph{AAAI},
  2020.

\bibitem{shi2020multi}
L.~Shi, S.~Geng, K.~Shuang, C.~Hori, S.~Liu, P.~Gao, and S.~Su, ``Multi-layer
  content interaction through quaternion product for visual question
  answering,'' \emph{arXiv preprint arXiv:2001.05840}, 2020.

\bibitem{huang2020aligned}
Q.~Huang, J.~Wei, Y.~Cai, C.~Zheng, J.~Chen, H.-f. Leung, and Q.~Li, ``Aligned
  dual channel graph convolutional network for visual question answering,'' in
  \emph{ACL}, 2020, pp. 7166--7176.

\bibitem{kervadec2020estimating}
C.~Kervadec, G.~Antipov, M.~Baccouche, and C.~Wolf, ``Estimating semantic
  structure for the vqa answer space,'' \emph{arXiv preprint arXiv:2006.05726},
  2020.

\end{thebibliography}
}
\end{small}
\vfill

\end{document}